# Predicting TUG score from gait characteristics with video analysis and machine learning

MA Jian

Hitachi (China) Research & Development Corporation
Email: *majian@hitachi.cn*

## Abstract

Fall is a leading cause of death which suffers the elderly and society. Timed Up and Go (TUG) test is a common tool for fall risk assessment. In this paper, we propose a method for predicting TUG score from gait characteristics extracted from video with computer vision and machine learning technologies. First, 3D pose is estimated from video captured with 2D and 3D cameras during human motion and then a group of gait characteristics are computed from 3D pose series. After that, copula entropy is used to select those characteristics which are mostly associated with TUG score. Finally, the selected characteristics are fed into the predictive models to predict TUG score. Experiments on real world data demonstrated the effectiveness of the proposed method. As a byproduct, the associations between TUG score and several gait characteristics are discovered, which laid the scientific foundation of the proposed method and make the predictive models such built interpretable to clinical users.

**Keywords**: copula entropy, gait characteristics, fall risk assessment, Timed Up and Go, linear regression, support vector regression

# 1. Introduction

Fall is common among the elderly, especially those living with Parkinson or Dementia. Fall injury usually leads to devastating consequences, especially for the elderly, from physical or mental impairments to lose of mobility and independence, even to death. According to the WHO report [1], fall was one of the top 20 leading causes of death in 2015, with about 714 thousands death worldwide accounting for 1.2% of mortality and was anticipated to remain on the top 20 list in 2030, with an estimated 976 thousands death or 1.4% of mortality. Besides life cost, fall injuries have also imposed economic cost upon our societies. For instance, elderly fall injuries cost about $20 billion per year in the United States alone [2].

Multiple risk factors were identified to highly likely lead to fall injuries, including muscle weakness, fall history, gait and balance deficits, use of assisting devices, visual deficits, arthritis, impaired activities of daily living, mental and cognitive impairments, and high age [3]. In this research, we will focus on an important fall risk factor -- gait impairments, with expectation of preventing fall through new technology for automatic fall risk assessment.

There are several available instruments for fall risk assessment, pertinently on gait and balance abilities, including Tinetti Performance Oriented Mobility Assessment (POMA), Berg Balance Test, Dynamic Gait Index, and Timed Up and Go (TUG) test [4]. Among them, TUG [6] are widely used due to ease to perform, low time cost, and reliable performance. TUG is a simple test assessing mobility by means of only five sequential tasks: rising from a chair, walking three meters, turning around, walking back to the chair, and sitting down [6]. However, in certain care settings, this instrument is still inconvenient for the elderly to perform many functional activities and time-consuming for caregivers as daily routine service. Meanwhile, TUG has also its limitations, including variability, measuring only simple tasks, sensitive to environmental conditions, subjective to professionals, etc [7]. Technology for monitoring and assessing senior’s fall risk, continuously and obtrusively, remains an appealing demand.

In this paper, we propose a new technology for automatic and unobtrusive TUG assessment with video analysis and machine learning techniques. With video analysis, gait characteristics will be extracted from video data. Their associations with fall risk scores will be measured with copula entropy and then the predictive models for predicting TUG score will be built on the mostly associated characteristics. Such technology is expected to has several advantages, such as monitoring senior’s functional condition automatically, continuously and unobtrusively. The model such developed is based on the nonlinear association between gait characteristics and fall risk which makes the predictive models interpretable to clinical users.

# 2. Related research

For the same issue, there are already several initial researches. In a pilot research, King et al [8] tried to use wearable sensors to assess fall risk and got some preliminary results on characteristics of sensor data from different fall risk groups. Another initial research by Rantz et al [9] studied correlation (linear associations) between six fall risk scores and gait characteristics (including stride time, stride length, and gait velocity) derived from Pulse-Doppler radar and Microsoft Kinect. Wearable inertial sensors were investigated to derive gait related parameters during TUG test to classify faller/non-faller [10][11]. More works on wearable inertial sensors for fall risk assessment were reviewed in [12][13].

Automatic TUG test based on different technologies, such as computer vision, depth camera, wearable sensors, or smart phone, are gaining momentum recently [7]. Li, et al proposed a automatic TUG sub-tasks segmentation method based on 2D camera [14]. The TUG test was divided into six sub-tasks. Pose was estimated from each video frame and the coordinates of pose were used as input of the classifiers to predict the sub-task of the frame. Depth camera, such as Kinect, has been applied to automatic TUG test by several researchers. Using skeleton or depth data, they all tried to estimate the TUG score by identifying six phases of the test. Savoie, et al [15] proposed a system for automating the TUG test using Kinect camera. 3D pose series were derived with the combination of 2D pose estimation and 3D depth information and then six sub-tasks were identified by detecting transition position. Kempel, et al [16] proposed a method for automatic TUG test with Kinect camera. Skeleton data and depth data of Kinect were utilized for detecting six phases of TUG test. Dubois, et al identified the phases of TUG test from the depth data of Kinect camera and additionally, extracted several gait related parameters to classify the subjects as with low or high fall risk [17].

Mehdizadeh, et al studied whether the gait characteristics extracted from Kinect vision system are associated with number of fall during two week [18]. The gait characteristics include five categories: spatial-temporal, variability, symmetry, stability, and acceleration frequency domain. Poisson regression was used to model the relationship between gait characteristics and number of falls. Only the characteristics selected with Pearson correlations are taken as input of the regression model.

# 3. Video analysis system

In this research, the fall prediction models are built upon video data obtained from the video analysis system illustrated in Figure 1 [5]. In the system, two type of video cameras, one 2D camera and one 3D camera (composed of two 2D cameras), are deployed in the living room of the elderly. All cameras work simultaneously. From 2D camera, video of human activities is recorded and then analyzed with pose estimation technique (Mask R-CNN [19] in our experiments), with which positions of human joints in the video frames are estimated. From 3D camera, depth information of each video frame is derived with stereo vision technique. Matching joint position and depth information, one can obtain 3D spatial information of human pose in every video frames. Such 3D information will be the raw data for the following ML pipeline to extract gait characteristics.

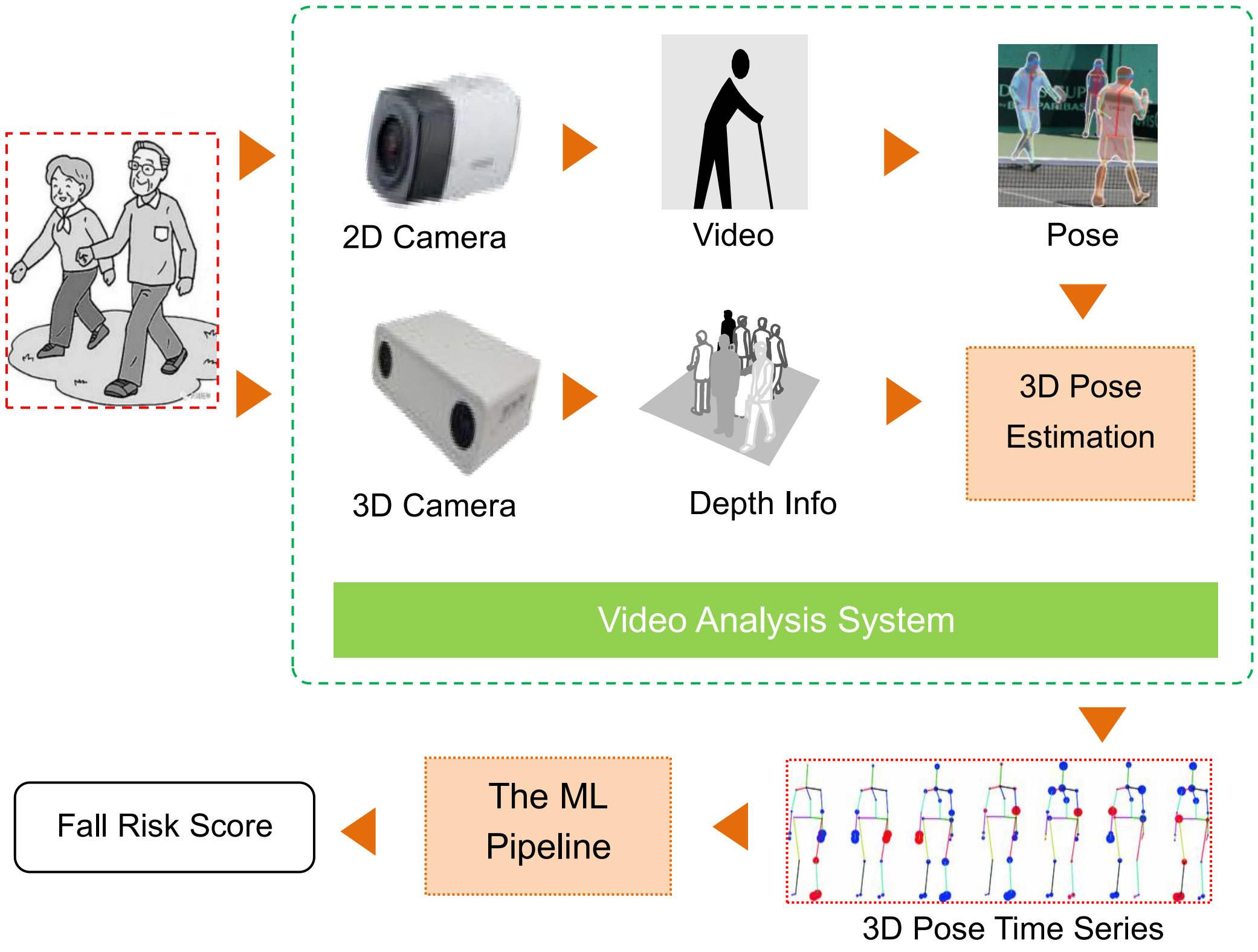

Figure 1. The video analysis system of the current research.

In this research, we are interested in gait characteristics associated with fall risk. 9 types of gait characteristics are considered to be extracted in our experiments (Table 1). All characteristics are computed from 3D pose time series of a short time interval of video. Note that 'Gait speed' and 'Step length' (or 'Pace') are two basic characteristics, and the other seven characteristics are derivatives of them. Specifically, we first generate a group of the characteristics from a series of short time interval of the video and then calculate the average and variance of the characteristics on all the intervals. In this way, each video will derive a sample data with 18 characteristics.

Table 1. Definitions of the gait characteristics extracted from video.

| **Characteristics** | **Definition** |
|---|---|
| Gait speed | Speed of body movement |
| Speed variability | standard deviation of stride speeds |
| Stride time | time between one peak and the second-next peak |
| Stride time variability | standard deviation of stride times |
| Stride frequency | median of modal frequency for the ML and half the modal frequenci es for the V and AP directions |
| Movement intensity | standard deviation of acceleration rate |
| Low-frequency percentage | Summed power up to a threshold frequency divided by total power |
| Acceleration range | Difference between minimum and maximum acceleration |
| Step length (Pace) | Length of one step |

# 4. Methodology

## 4.1 Copula Entropy

### 4.1.1 Theory

Copula theory unifies representation of multivariate dependence [20][21]. According to Sklar theorem [22], multivariate joint density function can be represented as a product of its marginals and copula density which represents dependence structure among random variables. Please refer to [23] for notations.

With copula density, one can define a new mathematical concept, called *Copula Entropy* [23], as follows:

**Definition** 1 (Copula Entropy). Let $\mathbf{X}$ be random variables with marginals $\mathbf{u}$ and copula density $c(\mathbf{u})$. Copula entropy of $\mathbf{X}$ is defined as

$$H_c(\mathbf{X}) = -\int_{\mathbf{u}} c(\mathbf{u}) \log c(\mathbf{u}) d\mathbf{u}. \tag{1}$$

In information theory, Mutual Information (MI) is a concept different from entropy [24]. In [23], Ma and Sun proved that MI is actually negative copula entropy, as follows:

**Theorem** 1. MI is equivalent to negative copula entropy:

$$I(\mathbf{X}) = -H_c(\mathbf{X}). \tag{2}$$

Theorem 1 has simple proof [23] and an instant corollary on the relationship between information containing in joint density function, marginals and copula density.

**Corollary** 1.

$$H(\mathbf{X}) = \sum_i H(X_i) + H_c(\mathbf{X}). \tag{3}$$

The above worthy-a-thousand-words results cast insight into the relation between MI and copula and therefore build the bridge between information theory and copula theory.

### 4.1.2 Estimation

MI, as a fundamental concept in information theory, has wide applications in physical, social, and biological sciences. However, estimating MI has been considered to be notoriously difficult. Under the blessing of Theorem 1, Ma and Sun [23] proposed a non-parametric method for estimating copula entropy (MI) from empirical data, which composes of only two simple steps:[1]

Step 1. estimating Empirical Copula Density (ECD);

Step 2. estimating copula entropy.

[1] The code is available at https://github.com/majianthu/copent.

For Step 1, if given data samples $\{\mathbf{x}_1,\ldots,\mathbf{x}_T\}$ i.i.d. generated from random variables $\mathbf{X}=[x_1,\ldots,x_N]^T$, one can easily derive ECD with empirical functions

$$F_i(x_i)=\frac{1}{T}\sum_{t=1}^{T}\chi(X_t^i\leq x_t^i), \quad (4)$$

where $i=1,\ldots,N$ and $\chi$ represents for indicator function. Let $\mathbf{u}=[F_1,\ldots,F_N]$, and then one derives a new samples set $\{\mathbf{u}_1,\ldots,\mathbf{u}_T\}$ as data from ECD $\mathrm{c}(\mathbf{u})$.

Once ECD is estimated, Step 2 is essentially a problem of entropy estimation which can be tackled by many existing methods. Among them, k-Nearest Neighbor method [25] was suggested in [23], which lead to a non-parametric way of estimating copula entropy.

## 4.2 Predictive models

### 4.2.1 Linear Regression (LR)

LR models linear relationship between dependent variable and many independent variables. Suppose there are dependent random variable Y and an independent random vector X, the LR model is as:

$$y = f(X). \quad (5)$$

The mathematical model for $f(\cdot)$ is linear functions whose parameters are estimated form sample data.

$$y=\mathbf{A}x+\varepsilon, \quad (6)$$

where A are parameters to be estimated, and $\varepsilon$ is error element.

The LR model is easy to estimate and interpretable, but unable to model nonlinear relation.

### 4.2.2 Support Vector Regression (SVR)

SVR is a popular ML method that can learn complex relationship from data [26][27]. Given a group of data and a hypothesis space, one can learn a cluster of function relationship. SVR can learn the model with the simplest possible model complexity and meanwhile do not compromise on prediction ability. This is due to max-margin principle.

Guided by the Max-Margin principle, the learning of SVR can be formulated as the following optimization problem [26]:

$$\min \frac{1}{2}\|w\|^2$$
$$s.t. \begin{cases} y_i - \langle w, x_i \rangle - b \le \varepsilon \\ \langle w, x_i \rangle + b - y_i \le \varepsilon \end{cases} \quad (7)$$

For notations please refer to [27]. This optimizing problem can be solved by quadratic programming techniques after transformed to its dual form. The original SVR problem is linear and can be upgraded into a nonlinear one by Kernel tricks. The final SVR model is as

$$f(x) = \sum_i v_i \, \mathrm{k}(x, x_i) + b \,, \quad (8)$$

where $x_i$ represents support vectors, and $k(\cdot,\cdot)$ represents kernel function. It can be learned that SVR is simply a linear regression model in nonlinear kernel space.

## 4.3 Pipeline of machine learning

We proposed a pipeline with the above concept and methods (Figure 2). The pipeline starts from collecting raw data, and then extracting a group of attributes from raw data. With copula entropy measuring association strength, one then selects the attributes most associated with TUG score. Such selected attributes are fed into the trained ML models (LR and SVR) to predict the scores. This pipeline has been applied to built cognitive assessment tool previously [35].

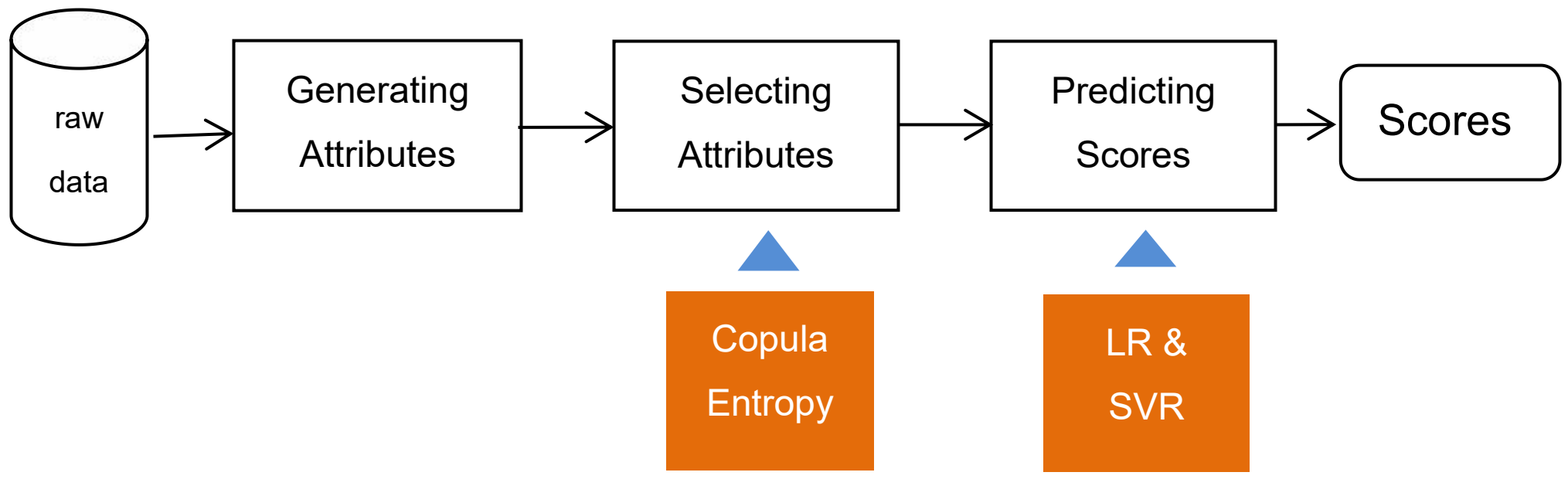

Figure 2. The pipeline of machine learning.

# 5. Data

The data for the experiments were collected from 40 subjects at Tianjin. All the participants signed informed consent. The subjects were administrated to perform TUG tests twice a day for several times in one month and did 146 tests totally. For each test, a video about 1-4 minutes was recorded and then gait characteristics were extracted from it with the above video analysis system [5]. With one sample composed of 18 characteristics for a video, a group of sample data were extracted from each video and then attached with the TUG scores corresponding to the video. In such way, the whole data set with 146 samples was finally generated from all the 146 videos.

The Distribution of TUG scores is shown in Figure 3. It can be learned that most tests were conducted on healthy subjects (TUG ≤ 10) and few subject was with high fall risk (TUG≥30).

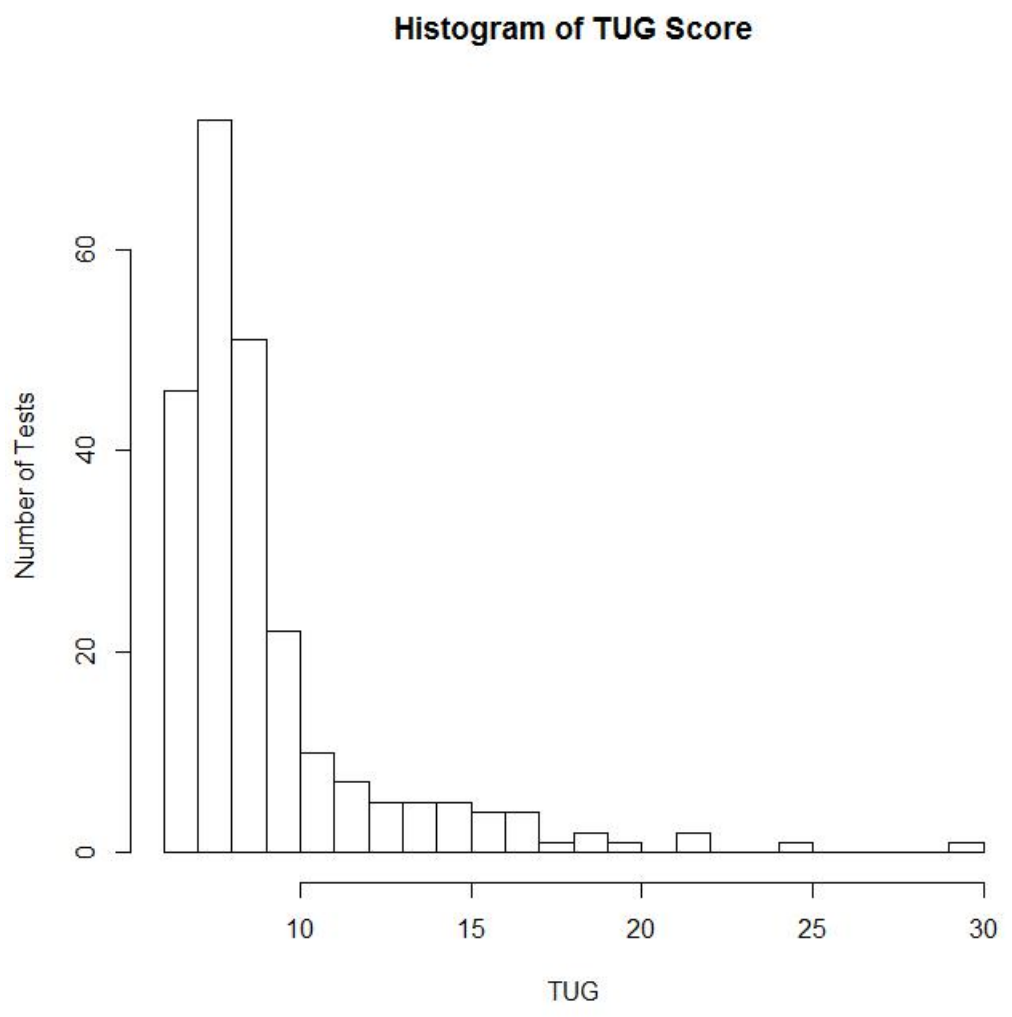

Figure 3. Distributions of the TUG scores in the experiments.

# 6. Experiments and results

## 6.1 Experiments

We conducted experiments to study the associations within data and the performance of the predictive models. The predictive models will be evaluated on two aspects: the interpretability of the characteristics and the prediction performance.

In the first experiment, the association between the TUG scores and the characteristics are measured with copula entropy. The most associated characteristics will be selected for the following prediction experiments.

In the second experiment, the selected characteristics will be used for predicting TUG score to check the performance of the predictive models. The predictive models in the experiments are Linear Regression (LR) and Support Vector Regression (SVR). The ratio between training data and test data are (80/20)% and the data set was randomly separated 100 times. The hyper-parameters of SVR are tuned to obtain the best possible prediction results.

The performance of the predictive models are measured from two aspects. It will be measured by Mean Absolute Error (MAE) between the true TUG score and the predicted ones. To check the clinical diagnosis ability of the predictive models, we take the clinical cutoff (TUG=13.5) to separate faller and non-faller. The diagnosis accuracy of the predictive models is derived by comparing the true diagnosis with the predicted diagnosis results.

## 6.2 Results

The association between the characteristics and the TUG score measured by copula entropy is shown in Figure 4. It can be learned from Figure 4 that 1) speed variance is the most associated characteristics with TUG score which is easy to understand and clinically meaningful; 2) the other characteristics with strong associations include gait speed, pace (step length), and acceleration range.

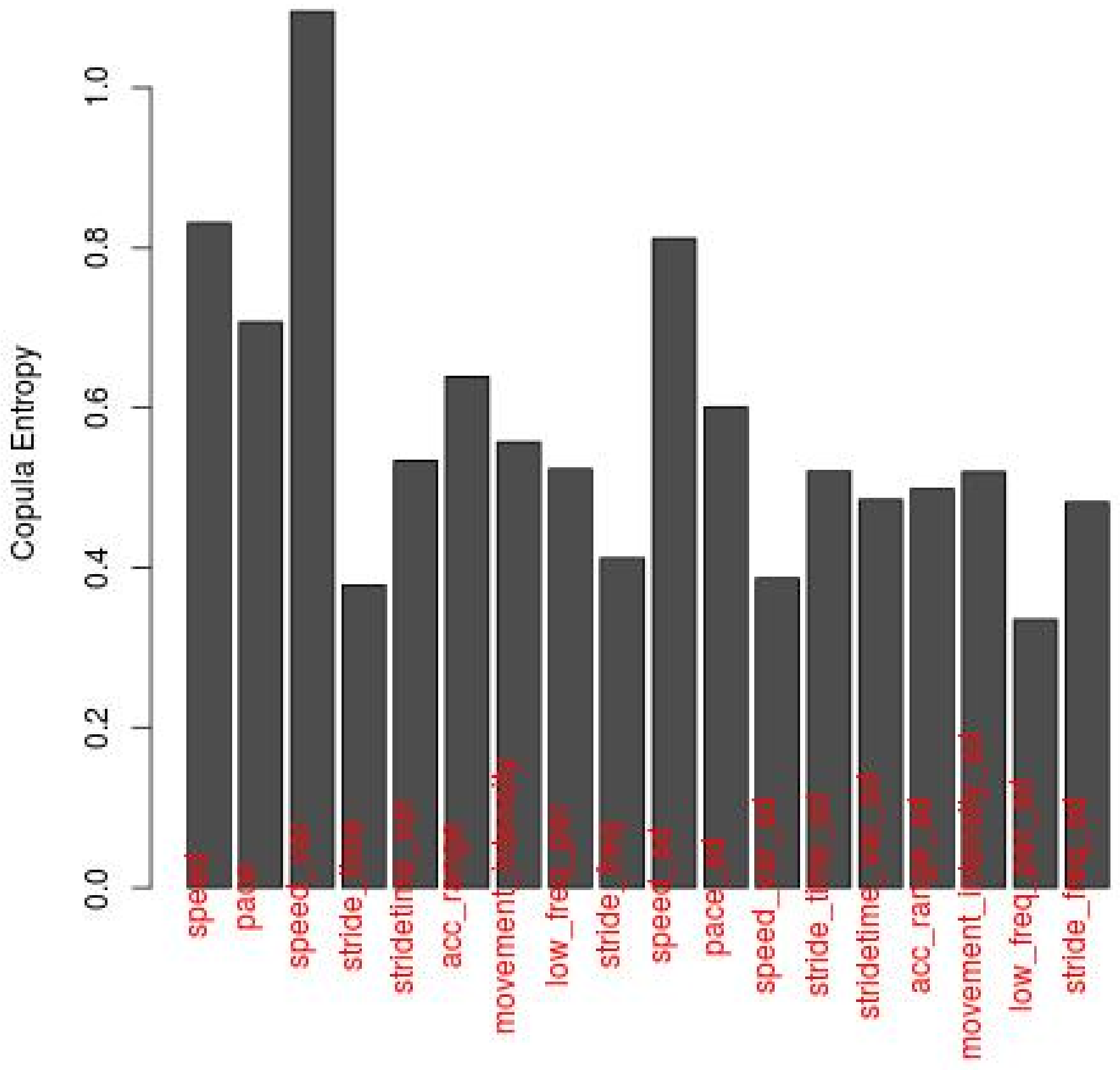

Figure 4. Association between the characteristics and the TUG score.

The joint distribution between TUG score and the three most associated characteristics are plotted in Figure 5, from which it can be learned that speed and pace are linearly associated and both non-linearly associated with TUG score. This may means that the method for generating gait characteristics is well defined and the characteristics are biologically and clinically plausible.

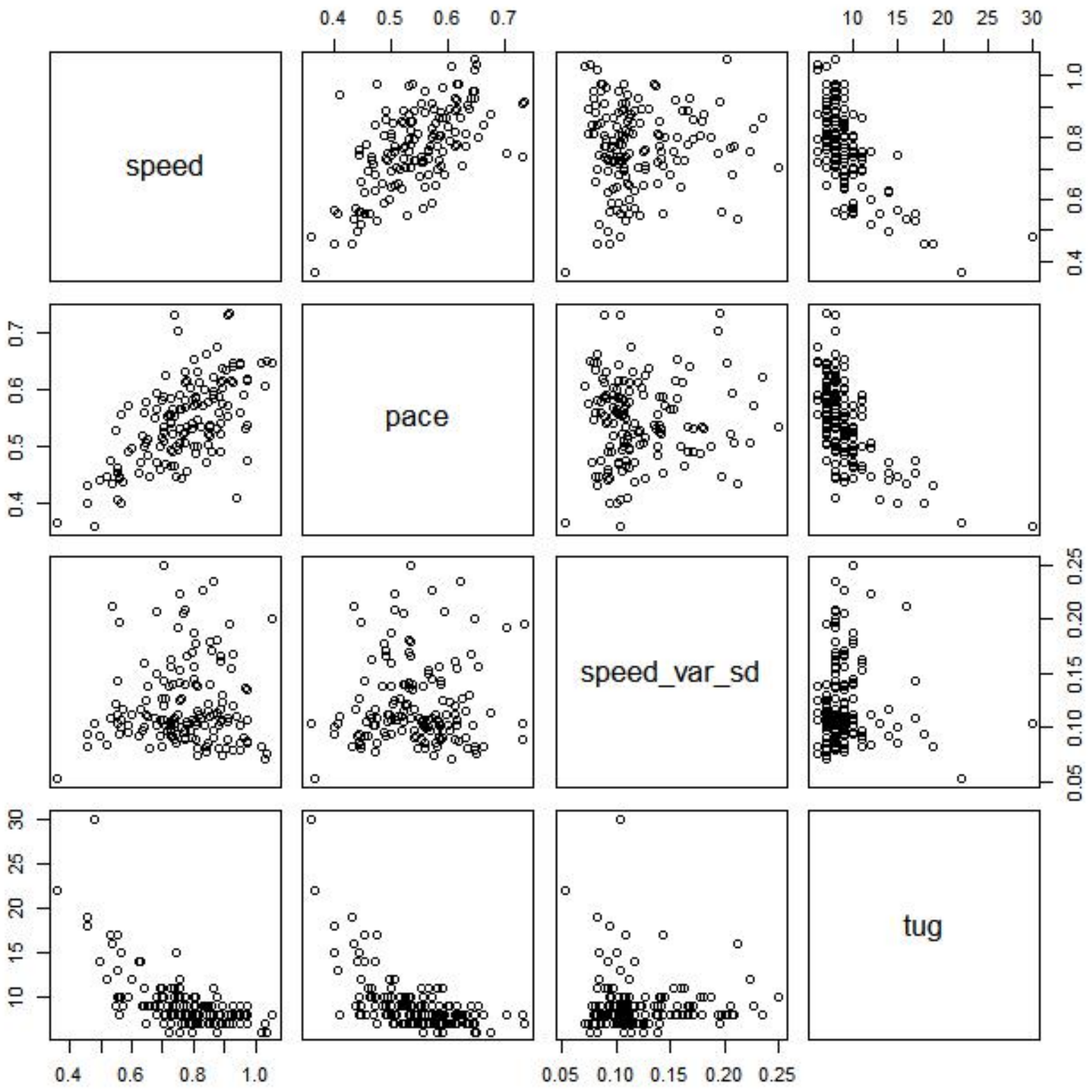

Figure 5. Joint distribution between the TUG score and the 3 characteristics.

Considering the associations between the characteristics and TUG score, two experiments on predicting TUG score are conducted. In the first experiment, three characteristics (gait speed, pace, and speed variance) are used as the input of the predictive models; In the second experiment, four characteristics, including gait speed, pace, speed variance, and acceleration range, are used as the input of the predictive models. The predicting results are shown in Figure 6 and 7. The performance of the two experiments in terms of MAE and diagnosis accuracy are listed in Table 2 and 3.

Comparing the results of two experiments between Table 2 and 3, one can learn that the performance of the predictive models are hardly improved by including another characteristics into models. This may imply that the three characteristics are mostly useful for prediction but including other characteristics are not helpful for prediction.

Comparing the prediction by LR and SVR between Table 2 and 3, one can learn that the latter is better in terms of MAE, and the former is better in terms of diagnosis accuracy. This is because that LR presents better results for the faller (with high TUG score) as shown in Figure 6 and 7.

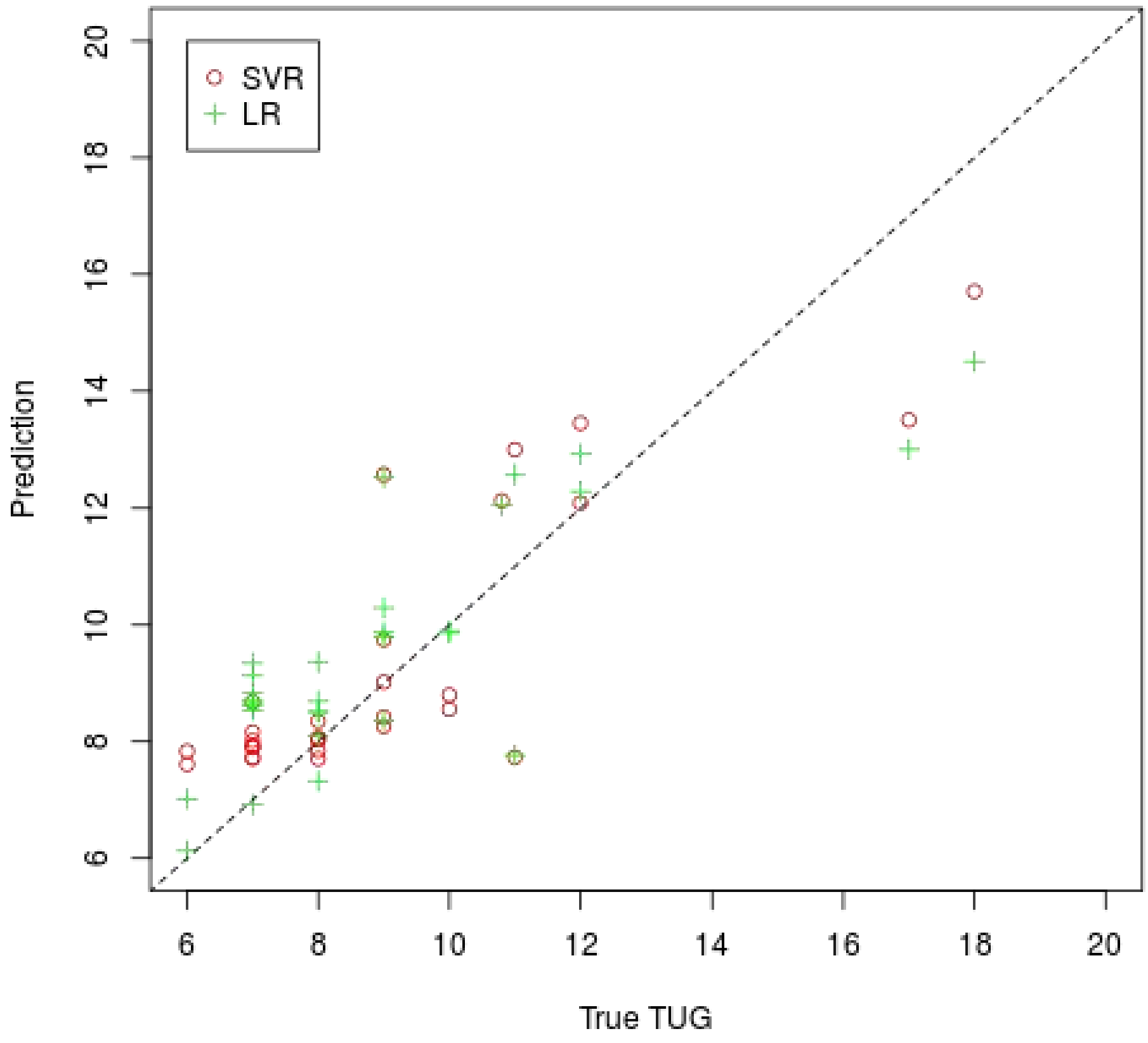

Figure 6. Prediction with the three characteristics.

Table 2. Performance of the predictive models with the 3 characteristics.

| | LR | SVR |
|---|---|---|
| MAE | 1.675 | 1.429 |
| Diagnosis Accuracy (%) | 94.4 | 92.7 |

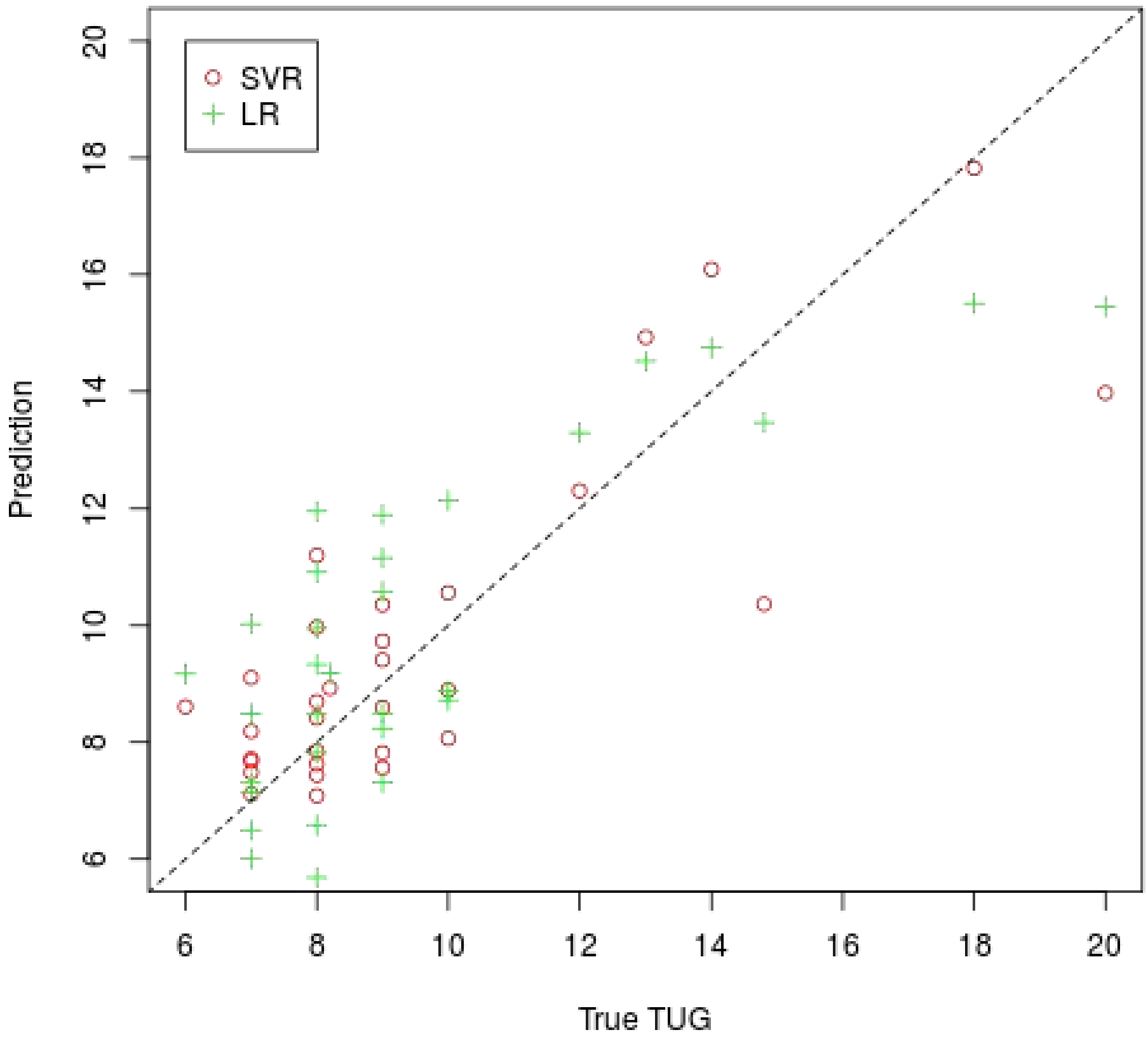

Figure 7. Prediction with the four characteristics.

Table 3. Performance of the predictive models with the 4 characteristics.

| | LR | SVR |
|---|---|---|
| MAE | 1.612 | 1.441 |
| Diagnosis Accuracy (%) | 95.4 | 93.2 |

# 7. Discussion

From Figure 4, one can learn that several characteristics, such as gait speed, pace, speed variance, and acceleration range are associated with TUG score. It is obvious that gait speed and pace are the mostly associated characteristics with TUG score. This indicates the method of generating characteristics is reasonable because gait speed has been widely considered as an effective indicator of functional ability in clinical research [28][29][30].

The other characteristics show also high association strength and contribute to the performance of the models, which may suggest that speed variance and acceleration range measure certain aspects of functional ability and is helpful for predicting TUG score. Previously, speed variance is suggested as a marker of fall risk [31]. Several studies also show that variance of gait speed may be much closely related to fall risk than the average of gait speed [32][33][34]. This point is supported by the evidence in our experiments that speed variance has much stronger association with TUG score than gait speed.

When examining the association pattern of gait speed, pace and TUG score, one can find that the associations are typical nonlinear with long tail in the distribution. This reflects how gait speed and pace change as functional ability deteriorates. This also implies that linear correlation coefficient is not a good choice for selecting characteristics for the predictive models.

Examining the results of two experiments carefully, one can learn that including more gait characteristics slightly improves the performance of both models. This implies that the two characteristics (gait speed and pace) are mostly informative for the prediction and that the other characteristics are also somewhat helpful. We believe that the performance of the models may be further improved if more reasonable characteristics related to fall risk are introduced into the models.

Though diagnosis accuracy of the predictive models are high, we do not consider this as a big success with caution. This is because that most samples are from healthy people and such imbalanced data makes the models trained from the data tending to predicting the subject as healthy people and hence majority of predictions are right. However, we can still notice that the models make good prediction for the samples from the patients with high fall risk, better than the ones in the previous research did.

Compared with the related works , our method is novel on two points. First, the subjects are not necessarily asked to perform TUG test even they did in our experiment. We need not try to identify the phases of TUG test as others did [14][15][16][17]. This makes the method automatic, unobtrusive and easy to deploy at any setting. Second, our method predicts TUG score instead of number of falls and meanwhile the prediction is based on the gait characteristics carefully selected with copula entropy instead of Pearson correlation coefficient [18]. This makes the method scientifically sound. It deserve a mention that there are already studies reporting the relationship of the characteristics selected, such as gait variability, to fall risk, as discussed above. Our research confirms this relationship instead of providing contradictory results as in [18].

# 8. Conclusion

In this paper, we study how to predict TUG score with gait characteristics extracted from video data. We propose a method that can extract gait characteristics from the whole-length video with pose estimation and stereo vision techniques. These gait characteristics are then selected by measuring the associations between the characteristics and TUG score with copula entropy and the selected characteristics are then fed into the predictive models (LR and SVR) to predict TUG score. Experiments on real world data show the effectiveness of the proposed method.

As a byproduct, the associations between TUG score and several gait characteristics, such as gait speed, pace, and gait speed variance, are discovered with copula entropy. This discovery provides another evidence of these gait characteristics as the markers of fall risk, and makes the predictive models interpretable, which is of critical importance to clinical users.

# Acknowledgement

The author thanks Zhang Pan for providing the data of gait characteristics.